\documentclass[runningheads]{llncs}
\usepackage{graphicx}
\usepackage{comment}
\usepackage{amsmath,amssymb,bm} 
\usepackage{color}
\usepackage{float}
\usepackage{subfigure}
\usepackage{amsmath}
\usepackage{graphics}
\usepackage[ruled]{algorithm2e}
\usepackage{xcolor}

\begin{document}
\pagestyle{headings}
\mainmatter
\def\ECCVSubNumber{3155}  

\titlerunning{ABS}
\title{Angle-based Search Space Shrinking for Neural Architecture Search} 

\authorrunning{Yiming Hu et al.}
\author{Yiming Hu\textsuperscript{1,3,}\thanks{Equal contribution. The work was done during the internship of Yiming Hu at
MEGVII Technology}, Yuding Liang\textsuperscript{2,$\star$}, Zichao Guo\textsuperscript{2,}\thanks{Corresponding author}, Ruosi Wan\textsuperscript{2}, Xiangyu Zhang\textsuperscript{2}, Yichen Wei\textsuperscript{2}, Qingyi Gu\textsuperscript{1}, Jian Sun\textsuperscript{2}}

\footnotetext[1]{This work is supported by The National Key Research and Development Program of China (No. 2017YFA0700800), Beijing Academy of Artificial Intelligence (BAAI) and the National Natural Science Foundation of China under Grants 61673376}

\institute{\textsuperscript{1}Institute of Automation, Chinese Academy of Sciences \textsuperscript{2}MEGVII Technology\\
\textsuperscript{3}School of Artificial Intelligence, University of Chinese Academy of Sciences \email{\{liangyuding,guozichao,wanruosi,zhangxiangyu,weiyichen,sunjian\}@megvii.com} \email{\{huyiming2016, qingyi.gu\}@ia.ac.cn}  }
\maketitle
\setcounter{figure}{0}

\begin{abstract}
In this work, we present a simple and general search space shrinking method, called Angle-Based search space Shrinking (ABS), for Neural Architecture Search (NAS). Our approach progressively simplifies the original search space by dropping unpromising candidates, thus can reduce difficulties for existing NAS methods to find superior architectures. In particular, we propose an angle-based metric to guide the shrinking process. We provide comprehensive evidences showing that, in weight-sharing supernet, the proposed metric is more stable and accurate than accuracy-based and magnitude-based metrics to predict the capability of child models. We also show that the angle-based metric can converge fast while training supernet, enabling us to get promising shrunk search spaces efficiently. ABS can easily apply to most of NAS approaches (e.g. SPOS, FairNAS, ProxylessNAS, DARTS and PDARTS). Comprehensive experiments show that ABS can dramatically enhance existing NAS approaches by providing a promising shrunk search space.

\keywords{angle, search space shrinking, NAS}
\end{abstract}

\section{Introduction}\label{sec:introduction}
Neural Architecture Search (NAS), the process of automatic model design has achieved significant progress in various computer vision tasks~\cite{zoph2018learning,chen2019detnas,liu2019auto,xu2019auto}. They usually search over a large search space covering billions of options to find the superior ones, which is time-consuming and challenging. Though many weight-sharing NAS methods~\cite{guo2019single,liu2018progressive,liu2018darts,cai2018proxylessnas,wu2019fbnet} have been proposed to relieve the search efficiency problem, the challenge brought by the large and complicated search space still remains. 

Shrinking search space seems to be a feasible solution to relieve the optimization and efficiency problem of NAS over large and complicated search spaces. In fact, recent studies~\cite{chen2019progressive,nayman2019xnas,noy2019asap,cai2019once,li2019improving} have adopted different shrinking methods to simplify the large search space dynamically. These methods either speed up the search process or reduce the optimization difficulty in training stage by progressively dropping unpromising candidate operators. Though existing shrinking methods have obtained decent results, it's still challenging to detect unpromising operators among lots of candidate ones. The key is to predict the capacity of candidates by an accurate metric. Existing NAS methods usually use accuracy-based metric~\cite{liu2018progressive,perez2018efficient,cai2019once,li2019improving} or magnitude-based metric~\cite{chen2019progressive,nayman2019xnas,noy2019asap} to guide the shrinking process. However, neither of them is satisfactory: the former one is unstable and unable to accurately predict the performance of candidates in weight-sharing
setting~\cite{zhang2020deeper}; while the later one entails the rich-get-richer problem~\cite{adam2019understanding,chen2019progressive}.

\begin{figure}[t]
\centering
 \includegraphics[width=0.9\linewidth]{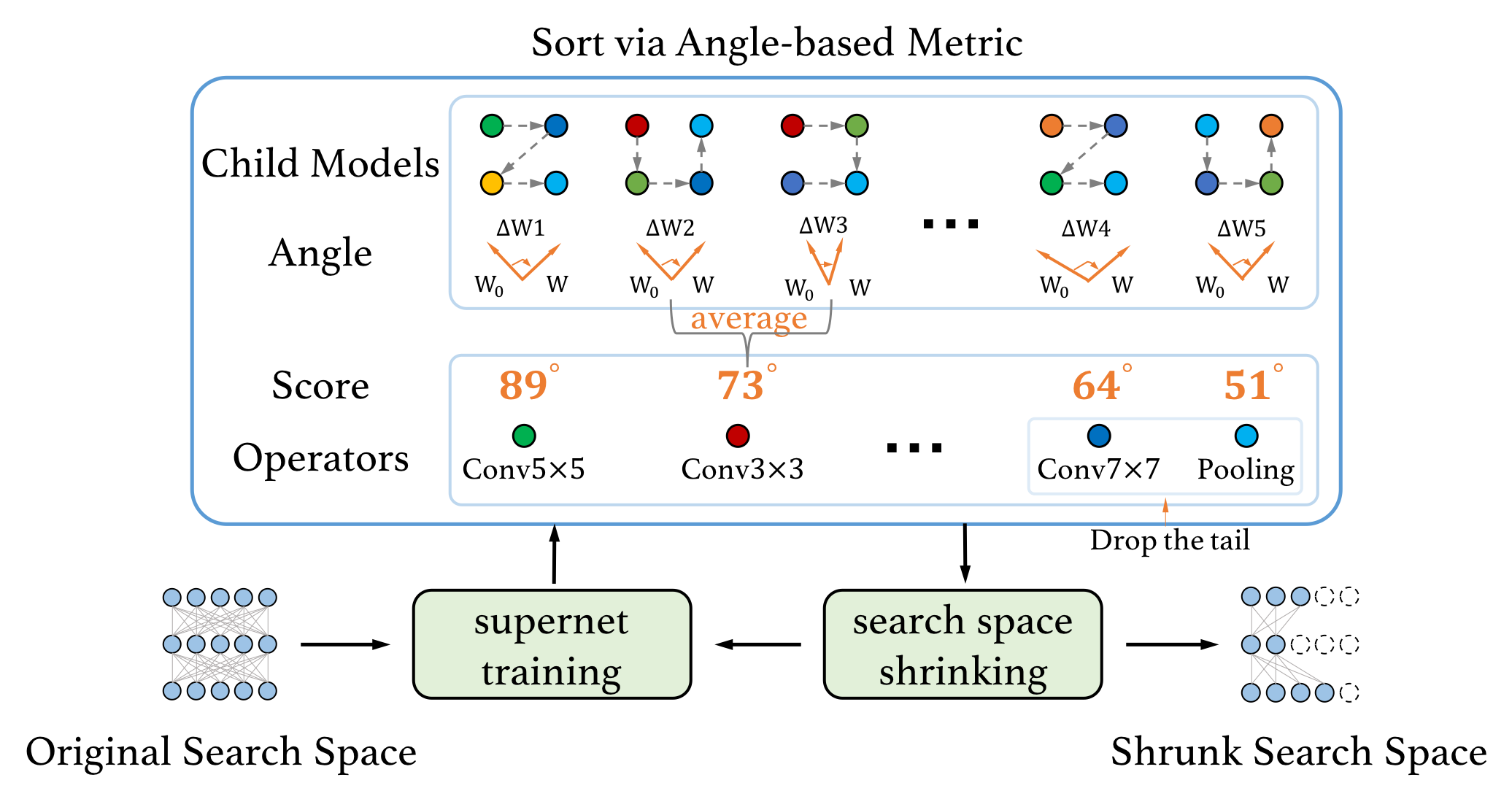}
\centering
\caption{Overview of the proposed angle-based search space shrinking method.
We first train the supernet for some epochs with uniform sampling. After this, all operators are ranked by their scores and those of them whose rankings fall at the tail are dropped
}
\label{fig:main_method}
\end{figure}

In this work, we propose a novel angle-based metric to guide the shrinking process. It's obtained via computing the angle between the model's weight vector and its initialization. Recent work \cite{carbonnelle2019layer} has used the similar metric to measure the generality of stand-alone models and demonstrates its effectiveness. For the first time, we introduce the angle-based metric to weight-sharing NAS. Compared with accuracy-based and magnitude-based metrics, the proposed angle-based metric is more effective and efficient. First, it can save heavy computation overhead by eliminating inference procedure. Second, it has higher stability and ranking correlation than accuracy-based metric in weight-sharing supernet. Third, it converges faster than its counterparts, which enables us to detect and remove unpromising candidates during early training stage.

Based on the angle-based metric, we further present a conceptually simple, flexible, and general method for search space shrinking, named as Angle-Based search space Shrinking (ABS). As shown in Fig.~\ref{fig:main_method}, we divide the pipeline of ABS into multiple stages and progressively discard unpromising candidates according to our angle-based metric. ABS aims to get a shrunk search space covering many promising network architectures. Contrast to existing shrinking methods, the shrunk search spaces ABS find don't rely on specific search algorithms, thus are available for different NAS approaches to get immediate accuracy improvement.

ABS applies to various NAS algorithms easily. We evaluate its effectiveness on Benchmark-201~\cite{dong2020nasbench201} and ImageNet~\cite{krizhevsky2017imagenet}. 
Our experiments show several NAS algorithms consistently discover more powerful architectures from the shrunk search spaces found by ABS. To sum up, our main contributions are as follows: 
\begin{enumerate}

\item We clarify and verify the effectiveness of elaborately shrunk search spaces to enhance the performance of existing NAS methods.
\item We design a novel angle-based metric to guide the process of search space shrinking, and verify its advantages including efficiency, stability, and fast convergence by lots of analysis experiments. 
\item We propose a dynamic search space shrinking method that can be considered as a general plug-in to improve various NAS algorithms including SPOS~\cite{guo2019single}, FairNAS~\cite{chu2019fairnas}, ProxylessNAS~\cite{cai2018proxylessnas}, DARTS~\cite{liu2018darts} and PDARTS~\cite{chen2019progressive}.

\end{enumerate}

\section{Related Work}\label{sec:related_work}
\subsubsection{Weight-sharing NAS.}
To reduce computation cost, many works~\cite{liu2018darts,cai2018proxylessnas,guo2019single,bender2018understanding,chen2019progressive} adopt weight-sharing mechanisms for efficient NAS. Latest approaches on efficient NAS fall into two categories: one-shot methods~\cite{bender2018understanding,guo2019single,chu2019fairnas} and gradient-based methods~\cite{liu2018darts,cai2018proxylessnas,wu2019fbnet}.
One-shot methods train an over-parameterized supernet based on various sample strategies~\cite{guo2019single,chu2019fairnas,bender2018understanding}.
After this, they evaluate many child models with the well-trained supernet as alternatives, and choose those with the best performance.
Gradient-based algorithms~\cite{liu2018darts,cai2018proxylessnas,wu2019fbnet} jointly optimize the network weights and architecture parameters by back-propagation. Finally, they choose operators by the magnitudes of architecture parameters.

\subsubsection{Search Space Shrinking.}
Several recent works~\cite{liu2018progressive,chen2019progressive,perez2018efficient,nayman2019xnas,noy2019asap,cai2019once,nayman2019xnas,li2019improving} perform search space shrinking for efficient NAS. For example, PDARTS~\cite{chen2019progressive} proposes to shrink the search space for reducing computational overhead when increasing network depth. In order to improve the ranking quality of candidate networks, PCNAS~\cite{li2019improving} attempts to drop unpromising operators layer by layer based on one-shot methods. However, existing shrinking techniques are strongly associated with specific algorithms, thus unable to easily apply to other NAS methods.
In contrast, our search space shrinking method is simple and general, which can be considered as a plug-in to enhance the performance of different NAS algorithms.
Moreover, an effective metric is vital to discover less promising models or operators for search space shrinking. Accuracy-based metric~\cite{liu2018progressive,perez2018efficient,cai2019once,li2019improving} and magnitude-based metric~\cite{chen2019progressive,nayman2019xnas,noy2019asap} are two widely used metrics in NAS area.  In contrast, our angle-based metric is much more stable and predictive without the poor ranking consistence and the rich-get-richer problem.

\subsubsection{Angle-based Metric.}
Recently, deep learning community comes to realize the angle of weights is very useful to measure the training behavior of neural networks: some works~\cite{li2019exponential,arora2018theoretical} theoretically prove that due to widely used normalization layers in neural network, the angle of weights is more accurate than euclidean distance to represent the update of weights; \cite{carbonnelle2019layer} uses the angle between the weights of a well-trained network and the initialized weights, to measure the generalization of the well-trained network on real data experiments. But the angle calculation method in \cite{carbonnelle2019layer} can't deal with none-parameter operators like identity and average pooling. To our best knowledge, no angle-based method was proposed before in NAS filed. Therefore we design a special strategy to apply the angle-based metric in NAS methods.

\section{Search Space Shrinking}

\label{sec:search_space_shrinking}
In this section, firstly we verify our claim that a elaborately shrunk search space can improve existing NAS algorithms by experiments. Then we propose an angle-based metric to guide the process of search space shrinking. Finally, we demonstrate the pipeline of the overall angle-based search space shrinking method.

\subsection{Elaborately Shrunk Search Space is Better}\label{sec:observation}
In this section, we investigate behaviors of NAS methods on various shrunk search spaces and point out that an elaborately shrunk search space can enhance existing NAS approaches. Our experiments are conducted on NAS-Bench-201~\cite{dong2020nasbench201}, which contains 15625 child models with ground-truths. We design 7 shrunk search spaces of various size on NAS-Bench-201, and evaluate five NAS algorithms~\cite{liu2018darts,dong2019one,dong2019searching,guo2019single,pham2018efficient} over shrunk search spaces plus the original one.

\begin{figure}[t]
\centering
 \includegraphics[width=0.8\linewidth]{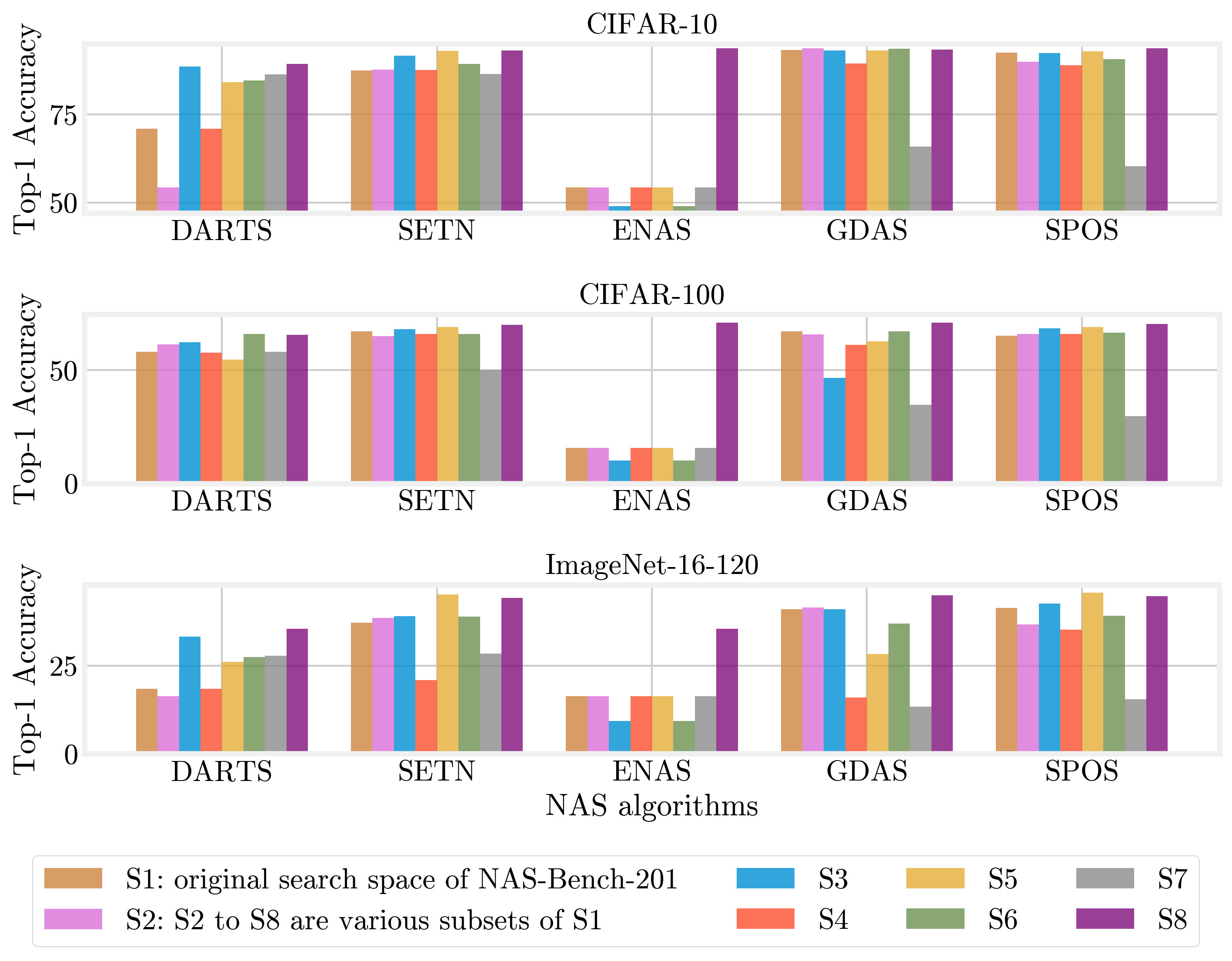}
\centering
\caption{Elaborately shrunk Search Space is better. We evaluate five different NAS algorithms~\cite{liu2018darts,dong2019one,dong2019searching,guo2019single,pham2018efficient} on eight search spaces
}
\label{fig:large_not_better}
\end{figure}

Fig. \ref{fig:large_not_better} summaries the experiment results. It shows the elaborately shrunk search space can improve the given NAS methods with a clear margin.
For example, GDAS finds the best model on CIFAR-10 from $S2$. On CIFAR-100 dataset, all algorithms discover the best networks from $S8$. For SPOS, the best networks found on ImageNet-16-120 are from $S5$. However, not all shrunk search spaces are beneficial to NAS algorithms. Most of shrunk search spaces show no superiority to the original one ($S1$), and some of them even get worse performance. Only a few shrunk search spaces can outperform the original ones, which makes it non-trivial to shrink search space wisely. 
To deal with such issue, we propose an angle-based shrinking method to discover the promising shrunk search space efficiently. The proposed shrinking procedure can apply to all existing NAS algorithms.
We'll demonstrate its procedure and effectiveness later. 

\subsection{Angle-based Metric}\label{angle_metric}
\subsubsection{Angle of Weights.}
According to \cite{li2019exponential,arora2018theoretical}, the weights of a neural network with Batch Normalization~\cite{ioffe2015batch,wan2020spherical} are ``scale invariant'', which means the Frobenius norm of weights can't affect the performance of the neural network and only direction of weights matters. Due to ``scale invariant" property, the angle $\Delta_{\bm{W}}$ (defined as Eq. \eqref{eq:angle}) between trained weights $\bm{W}$ and initialized weights $\bm{W}_0$ is better than euclidean distance of weights to represent the difference between initialized neural networks and trained ones:
\begin{equation}\label{eq:angle}
    \Delta_{\bm{W}} = arccos(\frac{<\bm{W}, \bm{W}_0 >}{||\bm{W}||_F \cdot ||\bm{W}_0||_F}),
\end{equation}
where $<\bm{W}, \bm{W}_0>$ denotes the inner product of $\bm{W}$ and $\bm{W}_0$, $||\cdot||_F$ denotes the Frobenius norm. \cite{carbonnelle2019layer} shows $\Delta_{\bm{W}}$ is an efficient metric to measure the generalization of a well-trained stand-alone model.

\subsubsection{Angle-based Metric for Child Model from Supernet.}
\begin{figure}[t]
\centering
\includegraphics[width=0.9\textwidth]{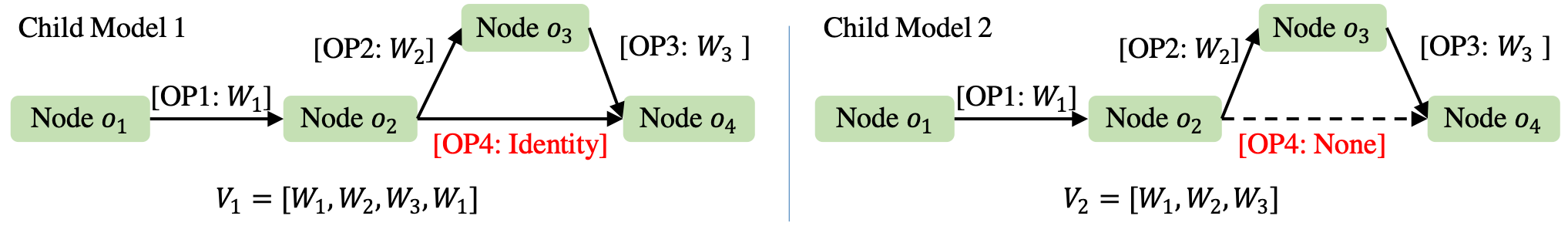}
\caption{Examples of the weight vector determined by structure and weights. $V_1, V_2$ are weight vectors of these child models respectively}
\label{fig: demmo_accum_angle}
\end{figure}

Since the angle shows close connection to generalization of trained networks, we consider using it to compare the performance of different child models. However, directly using angle $\Delta_{\bm{W}}$ of a child model may meet severe problems in weight sharing settings: \textbf{the procedure of computing $\Delta_{\bm{W}}$ can't distinguish different structures with exact same learnable weights}. Such dilemma is caused by non-parametric alternative operators (``none'', ``identity'', ``pooling''). For example, child model 1 and child model 2 shown in Fig. \ref{fig: demmo_accum_angle} have exact same learnable weights $[\bm{W}_1, \bm{W}_2, \bm{W}_3]$, but child model 1 has shortcut (OP4: identity), while child model 2 is sequential. Apparently child model 1 and 2 have different performance due to diverse structures, but $\Delta_{[\bm{W}_1, \bm{W}_2, \bm{W}_3]}$ can't reflect such difference. 

Therefore, to take non-parametric operators into account, we use the following strategy to distinguish different structures with the same learnable weights. For ``pooling'' and ``identity'' operators, we assign a fixed weight to them, and treat them like other operators with learnable weights: ``pooling''\footnotemark[2] has $k\times k$ kernel, where elements are all $1/k^2$, $k$ is the pooling size; ``identity'' has empty weights, which means we don't add anything to the weight vector for ``identity''. For ``none'' operator, it can totally change the connectivity of the child model, we can't simply treat it as the other operators. Hence we design a new angle-based metric as following to take connectivity of child model into account.

\footnotetext[2]{In this work, we do not distinguish ``max pooling'' and ``average pooling'' in our discussion and experiments.}

\paragraph{Definition of Angle-based Metric.}
\label{def:angle_based_metric}
Supernet is seen as a directed acyclic graph $\mathcal{G}(\bm{O}, \bm{E})$, where $\bm{O}=\{o_1, o_2, ..., o_M\}$ is the set of nodes, $o_1$ is the only root node (input of the supernet), $o_M$ is the only leaf node (output of the supernet); $\bm{E}=$\{$(o_i, o_j, w_{k})|$ alternative operators (including non-parametric operators except ``none'') from $o_i$ to $o_j$ with $w_{k}$\}. Assume a child model is sampled from the supernet, and it can be represented as a sub-graph $g(\bm{O}, \tilde{\bm{E}})$ from $\mathcal{G}$, where $\tilde{\bm{E}} \subset \bm{E}$, $o_1,o_M\in \tilde{\bm{E}}$; The angle-based metric $\Delta_{g}$ given $g$ is defined as:
\begin{equation}\label{eq:angle_based_metric}
    \Delta_{g} = arccos(\frac{<\bm{V}(g, \bm{W}_0), \bm{V}(g, \bm{W})>}{||\bm{V}(g, \bm{W}_0)||_F \cdot ||\bm{V}(g, \bm{W})||_F}),
\end{equation}
where $\bm{W}_0$ is the initialized weights of the supernet $\mathcal{G}$; $\bm{V}(g, \bm{W})$ denotes the weight vector of $g$, and it's constructed by concatenating the weights of all paths\footnotemark[3] from $o_1$ to $o_M$ in $g$, its construction procedure is shown in Algorithm \ref{alg:vec}. 

\footnotetext[3]{Path from node $o_{i_1}$ to node $o_{i_k}$ in a directed acyclic graph $\mathcal{G}(\bm{O}, \bm{E})$ means there exists a subset $P \subset \tilde{\bm{E}}$, where $P = \{(o_{i_1}, o_{i_2}, w_{j_1}), (o_{i_2}, o_{i_3}, w_{j_2}),..., (o_{i_{k-1}}, o_{i_k}, w_{j_{k-1}})\}$.}

The construction procedure described in Algorithm \ref{alg:vec} can make sure child models with diverse structures must have different weight vectors, even with the same learnable weights. As an example, Fig.\ref{fig: demmo_accum_angle} illustrates the difference between the weight vectors of child models with ``none'' and ``identity'' (comparing child model 1 and 2). Since $\bm{V}(g, \bm{W})$ is well defined on child models from any type of supernet, we compute the angel-based metric on all child models no matter whether there's ``none'' in supernet as an alternative.

\begin{algorithm}[t] \footnotesize
\caption{Construction of weight vector $\bm{V}(g, \bm{W})$ for Model $g$}
\label{alg:vec}
\LinesNumbered
\KwIn{A child model $g(\bm{O}, \tilde{\bm{E}})$ from the supernet, weights of supernet $\bm{W} = \{w\}$.}
\KwOut{weight vector $\bm{V}(g, \bm{W})$}
Find all paths from the root node $o_1$ to the leaf node $o_M$ in $g$: $\bm{P}=\{P \subset \tilde{\bm{E}} | \text{$P$ is a path from $o_1$ to $o_M$}\}$\;
$\bm{V} = [\emptyset]$($[\emptyset]$ means empty vector)\;
\For{$P$ in $\bm{P}$}{
    $\bm{V}_{P} = concatenate(\{w_k|(o_i, o_j, w_k) \in P\})$\;
    $\bm{V} = concatenate[\bm{V},\bm{V}_{P}]$\;
    }
$\bm{V}(g, \bm{W}) = \bm{V}$\;
\end{algorithm}

\subsubsection{Constructing Weight Vector on Cell-like/Block-like Supernet.}
Algorithm \ref{alg:vec} presents the general construction procedure of weight vector given a child model. It works well when the topology of supernet isn't too complex. However, in the worst case, the length of weight vector is of exponential complexity given the number of nodes. Hence it can make massive computational burden when number of nodes is too large in practice. Luckily, existing popular NAS search spaces all consist of several non-intersect cells, which allows us to compute the angle-based metric within each cell instead of the whole network. Specifically, we propose the following strategy as a computation-saving option:

\begin{enumerate}
\item Divide the whole network into several non-intersecting blocks;
\item Construct weight vector within each block respectively by Algorithm \ref{alg:vec};
\item Obtain weight vector of the child model by concatenating each block.
\end{enumerate}

\subsection{Angle-based Shrinking method}\label{method}
\subsubsection{Scores of Candidate Operators.}\label{score}
Before demonstrating the pipeline of angle-based shrinking method, we firstly define the angle-based score to evaluate alternative operators. Assume $\bm{P}=\{p_1, p_2, \cdots, p_N\}$ represents the collection of all candidate operators from supernet, $N$ is the number of candidate operators. We define the score of an operator by the expected angle-based metric of child models containing the operator:
\begin{equation}\label{eq:angle_score}
    Score(p_i) = \mathbb{E}_{g\in\{g|g \subset \mathcal{G}, \text{$g$ contains $p_i$}\}}\Delta_{g}, \quad i\in\{1,2,\cdots,N\},
\end{equation}
where $g$, $\mathcal{G}$ and $\Delta_{g}$ have been defined in section 3.2, $g$ is uniformly sampled from $\{g|g \subset \mathcal{G}, \text{$g$ contains $p_i$}\}$. In practice, rather than computing the expectation in Eq.\eqref{eq:angle_score} precisely, we randomly sample finite number of child models containing the operator, and use the sample mean of angle-based metric instead.

\subsubsection{Algorithm of Angle-based Shrinking Method.}
Based on the angle-based metric, we present Algorithm \ref{alg:ABS} to describe the pipeline shown in Fig.\ref{fig:main_method}. Note that during the shrinking process, at least one operator is preserved in each edge, since ABS should not change the connectivity of the supernet.

\begin{algorithm}[t] \footnotesize
\caption{Angle-based Search Space Shrinking Method (ABS)}
\label{alg:ABS}
\LinesNumbered
\KwIn{A supernet $\mathcal{G}$, threshold of search space size $\mathcal{T}$, number of operators dropped out per iteration $k$.}
\KwOut{A shrunk supernet $\tilde{\mathcal{G}}$}
Let $\tilde{\mathcal{G}}=\mathcal{G}$\;
\While{$|\tilde{\mathcal{G}}| >\mathcal{T} $}{
    Training the supernet $\tilde{\mathcal{G}}$ for several epochs following \cite{guo2019single}\;
    Computing score of each operator from $\tilde{\mathcal{G}}$ by Eq.\eqref{eq:angle_score}\;
    Removing $k$ operators from $\tilde{\mathcal{G}}$ with the lowest $k$ scores\;
    }
\end{algorithm}

\section{Experiments}\label{sec:experiment}

In this section, we demonstrate the power of ABS in two aspects: first, we conduct adequate experiments to verify and analyze the effectiveness of our angle-based metric in stability and convergence; second, we show that various NAS algorithms can achieve better performance by combining with ABS. 

\begin{table}[t]
\caption{The mean Kendall's Tau of 10 repeat experiments on NAS-Bench-201 for different initialization policies}
\label{table:tau_value_initialization}
\begin{center}
\begin{tabular}{cccc}
\hline
Initialization & CIFAR-10 & CIFAR-100 & ImageNet-16-120\\
\hline
    Kaiming-norm \cite{he2015delving} & $0.622$ & $0.608$ & $0.534$\\
    Xavier-uniform \cite{glorot2010understanding} & $0.609$ & $0.614$ & $0.544$\\
    Orthogonal & $0.609$ & $0.612$ & $0.533$ \\
\hline
\end{tabular}
\end{center}
\end{table}

\subsection{Empirical Study on Angle-based Metric}

\subsubsection{How important is the specific network initialization?} 
There are several different ways to initialize a network, while almost all initialization methods are gaussian type, thus the direction of initialized weights is always uniformly random. Theoretically, various initialization methods make no difference to the angle-based metric. The results in Table \ref{table:tau_value_initialization} prove our justification. Our proposed metric is reasonably robust to various initialization settings on different datasets. 

\subsubsection{Ranking Correlation in Stand-alone Model.}
First of all, we conduct experiments to verify if the angle-based metric defined in Eq.\eqref{eq:angle_based_metric} can really reflect the capability of stand-alone models with different structures. In detail, we uniformly select 50 child models from NAS-Bench-201 and train them from scratch to obtain the fully optimized weights. Since the initialized weights are known, the angle of a model can be calculated as Eq.\eqref{eq:angle_based_metric}. To quantify the correlation between the networks' capability and their angles, we rank chosen 50 models according to their angle, and compute the Kendall rank correlation coefficient~\cite{kendall1938new} (Kendall's Tau for short) with their accuracy in standalone setting. 

\begin{figure}[t]
\centering
 \includegraphics[width=0.8\linewidth]{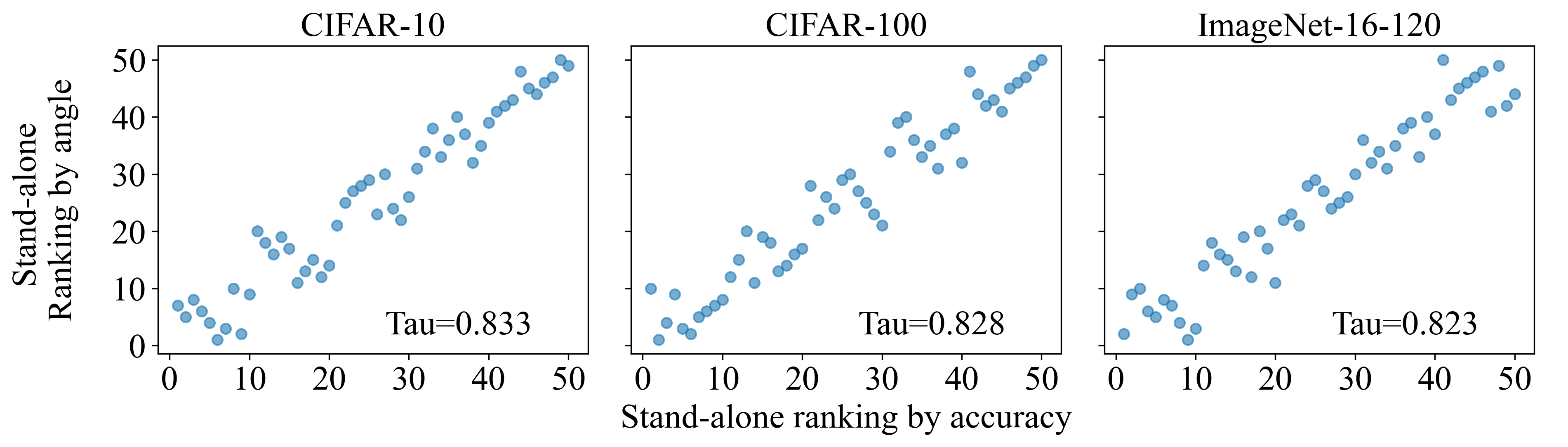}
\centering
\caption{
The correlation between the angle-based ranking and ground-truth ranking. We uniformly choose 50 models from NAS-Bench-201~\cite{dong2020nasbench201}, and train them from scratch. After this, we leverage the angle and accuracy of these models to rank them respectively. 
}
\label{fig:network_ranking}

\end{figure}

Fig.\ref{fig:network_ranking} shows the correlation between the network ranking by angle and the ranking of ground-truth on three datasets (CIFAR-10, CIFAR-100, ImageNet-16-120). It's obvious that the Kendall's Tau on all three different datasets are greater than 0.8, which suggests the angle of a model has significant linear correlation with its capability. Therefore, it's reasonable to use the angle-based metric to compare the performance of trained models even with different structures.

\subsubsection{Ranking Correlation in Weight-sharing Supernet.}
In this section, we verify the effectiveness of our angel-based metric in weight-sharing supernet.  In detail, we firstly train a weight-sharing supernet constructed on NAS-Bench-201 following~\cite{guo2019single}. Then we calculate different metrics, such as accuracy and angle, of all child models by inheriting optimized weights from supernet. At last, we rank child models by the metric and ground-truth respectively, and compute the Kendall's Tau between these two types of ranks as the ranking correlation. Since the magnitude-based metric can only rank operators, it's not compared here.

\begin{table}[t] 
\caption{The mean Kendall's Tau of 10 repeat experiments on NAS-Bench-201}
\label{table:tau_value}
\begin{center}
\begin{tabular}{ccccc}
\hline
Method & CIFAR-10 & CIFAR-100 & ImageNet-16-120\\
\hline
    Random & $0.0022$ & $-0.0019$ & $-0.0014$\\
    Acc. w/ Re-BN\footnotemark[4]~\cite{guo2019single} & $0.5436$ & $0.5329$ & $0.5391$\\
    Angle & $0.5748$ & $0.6040$ & $0.5445$ \\
\hline
\end{tabular}
\end{center}
\end{table}
\footnotetext[4]{\emph{Re-BN} means that before inferring the selected child model, we reset the batch normalization's~\cite{ioffe2015batch} mean and variance and re-calculate them on the training dataset.}

Table \ref{table:tau_value} shows the ranking correlations based on three metrics (random, accuracy with Re-BN, angle-based metric) on three different datasets (CIFAR-10, CIFAR-100, ImageNet-16-120). Accuracy-based metric with Re-BN and angle-based metric are both dramatically better than random selection. Importantly, our angle-based metric outperforms accuracy-based metric with a clear margin on all three datasets, which suggests that our angle-based metric is more effective to evaluate the capability of child models from supernet.

\begin{figure}[t]
\centering
 \includegraphics[width=0.8\linewidth]{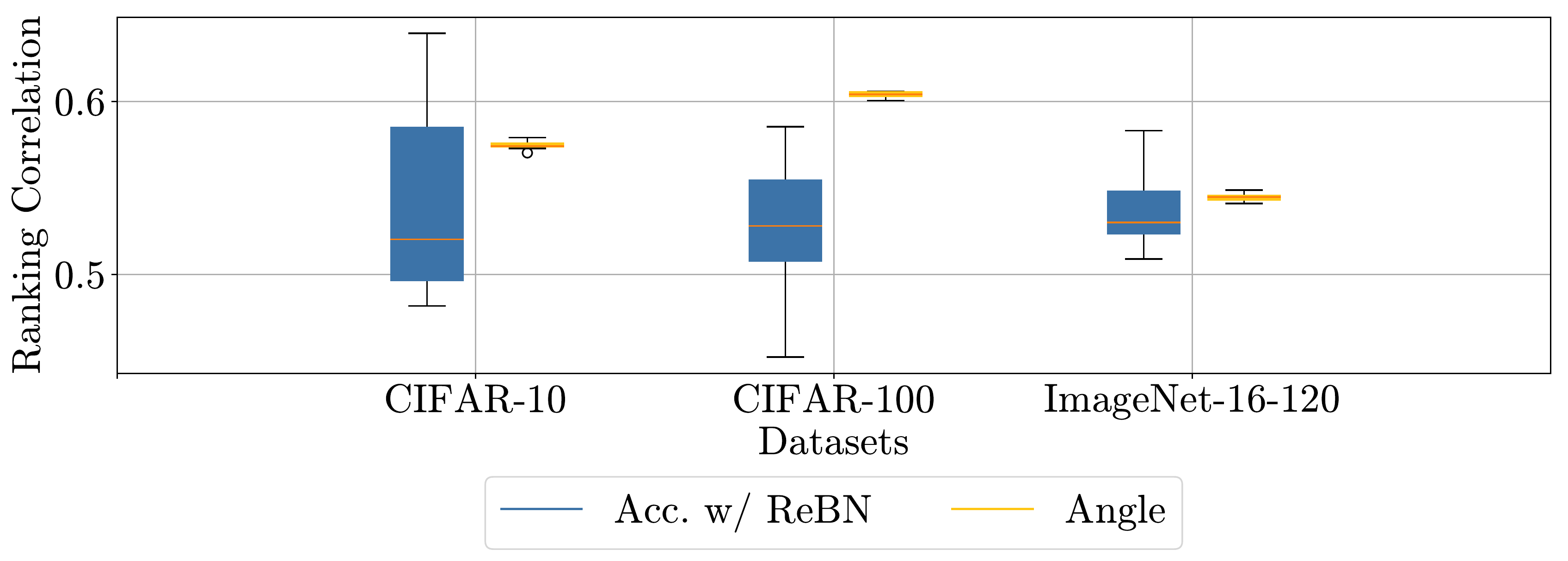}
\centering
\caption{The ranking stability on NAS-Bench-201. Every column is the range of ranking correlation for a metric and dataset pair. The smaller column means more stable}
\label{fig:different_running}
\end{figure}

\subsubsection{Ranking Stability.}
\label{sec:ranking_stability} 
We have shown that our angle-based metric can achieve higher ranking correlation than the accuracy-based metric. In this section, we further discuss the ranking stability of our metric. In detail, we conduct 9 experiments on three different datasets and calculate means and variances of ranking correlation obtained by accuracy-based metric and angle-based metric.  As Fig.\ref{fig:different_running} shows, our angle-based metric is extremely stable compared with accuracy-based metric. It has much smaller variance and higher mean on all three datasets. This is a crucial advantage for NAS methods, which can relieve the problem of reproducibility in weight-sharing NAS approaches. Magnitude-based metric is still not included in discussion, because it can't rank child models.

\begin{figure}[t]
\centering
 \includegraphics[width=0.8\linewidth]{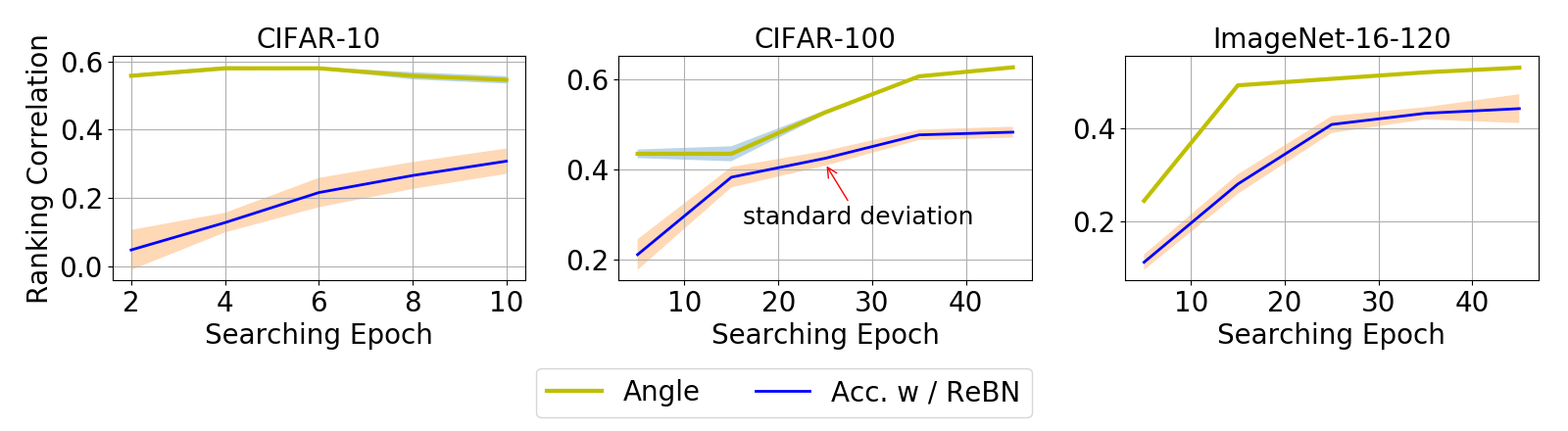}
\centering
\caption{Ranking correlation of metrics at early training stage on NAS-Bench-201}
\label{fig:spos_rank_correlation}
\end{figure}

\subsubsection{Convergence in Supernet Training.} 
In this section, we further investigate convergence behaviors of angle-based metric and accuracy-based metric in supernet training. In search space shrinking procedure, unpromising operators are usually removed when supernet isn't well trained. Hence the performance of the metric to evaluate child models' capability at early training stage will severely influence the final results. Fig. \ref{fig:spos_rank_correlation} shows different metrics' ranking correlation with ground-truth at early training stage. Our angle-based metric has higher ranking correlation on all three datasets during the first 10 epochs. Especially, there is a huge gap between such two metrics during the first 5 epochs. It suggests that our metric converges faster than accuracy-based metric in supernet training, which makes it more powerful to guide shrinking procedure at early training stage.

\subsubsection{Time Cost for Metric Calculation.}
Magnitude-based metric needs to train extra architecture parameters as metric except network weights, which costs nearly double time for supernet training. Instead, accuracy-based metric only requires inference time by inheriting weights from supernet. But it still costs much time when evaluating a large number of child models. And our angle-based metric can further save the inference time. To compare the time cost of calculating metrics, we train a supernet and apply the specific metric to 100 randomly selected models from NAS-Bench-201. Experiments are run ten times on \emph{NVIDIA GTX 2080Ti} GPU to calculate the mean and standard deviation. From Table \ref{table:processing_time}, the time cost of our metric on three datasets are all less than 1 seconds while accuracy-based metric's time cost are greater than 250 seconds.

\begin{table}[t] 
\caption{The processing time (100 models) of different metrics on NAS-Bench-201}
\label{table:processing_time}
\begin{center}
\begin{tabular}{ccccc}
\hline
Method & CIFAR-10 (s) & CIFAR-100 (s) & ImageNet-16-120 (s)\\
\hline
    Acc. w/ Re-BN~\cite{guo2019single} & $561.75_{\pm126.58}$ & $332.43_{\pm59.18}$ & $259.84_{\pm31.90}$\\
    Angle & $0.92_{\pm0.06}$ & $0.77_{\pm0.02}$ & $0.73_{\pm0.04}$ \\
\hline
\end{tabular}
\end{center}
\end{table}

\subsubsection{Select Promising Operators.}
The experiments above prove the superiority of angle-based metric as an indicator to evaluate child models from supernet, but we still need to verify if it's really helpful to guide the selection of promising operators. To this end, we directly compare the shrinking results based on different metrics. In our settings, the ground-truth score of each operator is obtained by averaging the ground-truth accuracy of all child models containing the given operator, the ground-truth ranking is based on the ground-truth score. We also rank alternative operators according to their metric-based scores: our angle-based score is defined as Eq.\eqref{eq:angle_score}; the accuracy-based score is similar to the ground-truth score but the accuracy is obtained from the well-trained supernet. It shares the same algorithm pipeline (see Algorithm \ref{alg:ABS}) and the hyper-parameters as our approach except the specific metric; the magnitude-based metric takes the magnitudes of corresponding architecture parameters as operator scores. It trains supernet following \cite{wang2019scalable}, but has the identical  training and shrinking setting as our method. After getting the metric based rank, we drop 20 operators with the lowest ranking, and check the ground-truth ranking of the reserved operators.

\begin{figure}[t]
	\centering
	\includegraphics[width=0.8\linewidth]{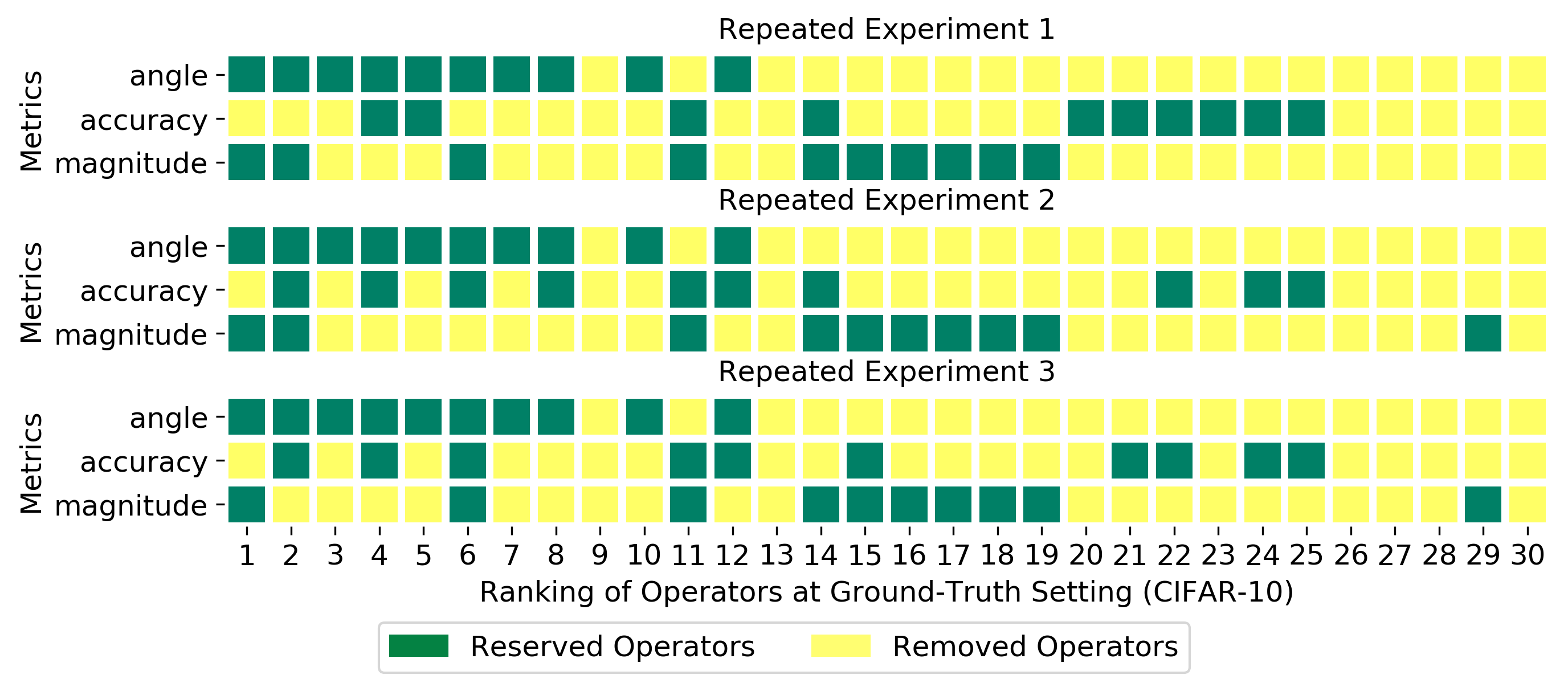}
	\centering
	\caption{The operator distribution after shrinking in three repeated CIFAR-10 experiments on NAS-Bench-201 with different random seeds
		}
	\label{fig:lefr_removed_operators}
\end{figure}

From Fig.\ref{fig:lefr_removed_operators}, magnitude-based and accuracy-based metrics both remove most of operators in the first 8 ground-truth ranking, while the angle-based metric reserves all of them. Moreover, almost all reserved operators our approach finds have higher ground-truth scores than the removed ones', while other two methods seem to choose operators randomly. Besides, we repeat the experiments for three times with different random seeds, the result shows that angle-based shrinking method can stably select the promising operators with top ground-truth scores, while the shrunk spaces based on the other metrics are of great uncertainty.

Though there's no guarantee that the child models with best performance must be hidden in the shrunk search space, it's reasonable to believe we are more likely to discover well-behaved structures from elaborately shrunk search spaces with high ground-truth scores. Based on this motivation, the angle-based metric allows us to select those operators with high performance efficiently.

\subsection{NAS Algorithms with Angle-based Shrinking}
In this part, we verify the power of ABS combined with existing NAS algorithms. We choose five NAS algorithms (SPOS\cite{guo2019single}, FairNAS\cite{chu2019fairnas}, ProxylessNAS\cite{cai2018proxylessnas}, DARTS\cite{liu2018darts}, and PDARTS\cite{chen2019progressive}), whose public codes are available, to apply ABS. All experiments are conducted on ImageNet. The original training set is split into two parts: 50000 images for validation and the rest for training.

\subsubsection{MobileNet-like Search Space.}
MobileNet-like Search Space consists of MobileNetV2 blocks with kernel size $\{3, 5, 7\}$, expansion ratio $\{3, 6\}$ and identity as alternative operators. We test the performance of ABS with SPOS~\cite{guo2019single}, ProxylessNAS~\cite{cai2018proxylessnas} and FairNAS~\cite{chu2019fairnas} on MobileNet-like search space.
SPOS and ProxylessNAS are applied on the Proxyless (GPU) search space~\cite{cai2018proxylessnas}, while FairNAS is applied on the same search space as~\cite{chu2019fairnas}.
We shrink the MobileNet-like search spaces by ABS at first, then apply three NAS algorithms to the shrunk spaces. 

In detail, supernet is trained for 100 and 5 epochs in the first and other shrinking stages. We follow the block-like weight vector construction procedure to compute the angle-based metric. The score of each operator is acquired by averaging the angle of $1000$ child models containing the given operator. Moreover, the base weight $\bm{W}_0$ used to compute angle is reset when over 50 operators are removed from the original search space. Because our exploratory experiments (see Fig. 4 in appendix) show that after training models for several epochs, the angle between the current weight $\bm{W}$ and the initialized weight $\bm{W}_0$ is always close to $90^{\circ}$ due to very high dimension of weights. It doesn't mean the training is close to be finished, but the change of angle is too tiny to distinguish the change of weights. Therefore, to represent the change of weights effectively during the mid-term of training, we need to reset the base weight to compute the angle.

When sampling child models, ABS dumps models that don’t satisfy flops constraint. For SPOS and ProxylessNAS, ABS removes $7$ operators whose rankings fall at the tail. For FairNAS, ABS removes one operator for each layer each time because of its fair constraint. The shrinking process finishes when the size of search space is less than $10^5$. For re-training, we use the same training setting as \cite{guo2019single} to retrain all the searched models from scratch, with an exception: dropout is added before the final fully-connected and the dropout rate is $0.2$. 

\begin{table}[t] 
\caption{Search results on MobileNet-like search space. $^*$ The searched models in their papers are retrained using our training setting}
\label{table:search_result_mobilenet}
\begin{center}
\begin{tabular}{ccccc}
\hline
 & Flops & Top1 Acc. & Flops(ABS) & Top1 Acc.(ABS)\\
\hline
    FairNAS~\cite{chu2019fairnas} & 322M & 74.24\%$^*$ & 325M & \textbf{74.42\%}\\
    SPOS~\cite{guo2019single} & 465M & 75.33\%$^*$ & 472M & \textbf{75.97\%} \\
    ProxylessNAS~\cite{cai2018proxylessnas} & 467M & 75.56\%$^*$ & 470M & \textbf{76.14\%} \\
\hline
\end{tabular}
\end{center}
\end{table}

As Table \ref{table:search_result_mobilenet} shows, all algorithms can obtain significant benefits with ABS. SPOS and ProxylessNAS find models from the shrunk search space with $0.6\%$ higher accuracy than from the original search space respectively. FairNAS also finds better model on shrunk search space with $0.2\%$ accuracy improvement.

\subsubsection{DARTS Search Space.}
Following the experiment settings in \cite{liu2018darts}, we apply the search procedure on CIFAR-10, then retrain the selected models from scratch and evaluate them on ImageNet.  In detail, the block-like weight vector construction procedure is adopted while using ABS. Supernet is trained for 150 and 20 epochs in the first and other shrinking stages respectively. More epochs are adopted for training supernet due to its slow convergence on DARTS search space. ABS removes one operator for each edge each time. The shrinking process stops when the size of shrunk search space is less than a threshold. DARTS and PDARTS share the same threshold as the MobileNet-like search space. During re-training, all algorithms use the same training setting as \cite{chen2019progressive} to retrain the searched models.

\begin{table}[t]
\caption{ImageNet results on DARTS search space. $^*$ For the form x(y), x means models searched by us using codes, y means the searched models in their papers}
\label{table:search_result_darts_imagenet}
\begin{center}
\begin{tabular}{cccc}
\hline
Method & Channels & Flops & Top1 Acc.\\
\hline
    DARTS~\cite{liu2018darts} & $48(48)^*$ & 446M(530M)$^*$ & $73.39\%(74.88\%)^*$\\
    DARTS(ABS) & $48$ & 619M & \textbf{75.59}\\
    DARTS(ABS, scale down) & $45$ & 547M & \textbf{75.19}\\
    PDARTS~\cite{chen2019progressive} & $48(48)^*$ & 564M(553M)$^*$ & $75.02\%(75.58\%)^*$\\
    PDARTS(ABS) & $48$ & 645M & \textbf{75.89}\\
    PDARTS(ABS, scale down) & $45$ & 570M & \textbf{75.64}\\
\hline
\end{tabular}
\end{center}
\end{table}

From Table \ref{table:search_result_darts_imagenet}, the architectures found by DARTS and PDARTS with ABS on CIFAR10 perform well on ImageNet. Equipped with ABS, DARTS and PDARTS get $2.2\%$ and $0.87\%$ accuracy improvement respectively without any human interference ($0.71\%$ and $0.31\%$ improvement even compared with reported results in \cite{liu2018darts,chen2019progressive}). Such vast improvement is probably due to the fact that the architectures found from the  shrunk search space have more flops. But it's reasonable that models with higher flops are more likely to have better capability if the flops are not constrained. Furthermore, to fairly compare the performance with constrained flops, the channels of the architecture we found from shrunk space are scaled down to fit with the constraint of flops. Table \ref{table:search_result_darts_imagenet} shows that even the constrained models from the shrunk search space can still get better results.

\subsubsection{Discussion.} 
Search space shrinking is very useful for NAS \cite{chen2019progressive,li2019improving}, and the angle-based metric is extremely suitable for shrinking due to its high correlation with the performance of DNN and fast convergence (see Fig. \ref{fig:spos_rank_correlation}). Our results show ABS can enhance existing NAS algorithms (see Table \ref{table:search_result_mobilenet}, \ref{table:search_result_darts_imagenet}). But the metric is not a perfect indicator (see Table \ref{table:tau_value}), so directly searching with our metric shows no superiority to combining it and other NAS methods: on the Mobilenet-like search space, our experiments indicate SPOS gets only 0.19\% improvement by replacing the accuracy-based metric with the metric, while combining with ABS, SPOS can get 0.64\% improvement. Thus we leverage the metric to perform shrinking.

\section{Conclusion and Future Work}\label{sec:conclusion}

In this paper, we point out that elaborately shrunk search space can improve performance of existing NAS algorithms. Based on this observation, we propose an angle-based search space shrinking method available for existing NAS algorithms, named as ABS. While applying ABS, we adopt a novel angle-based metric to evaluate capability of child models and guide the shrinking procedure. We verify the effectiveness of the angle-based metric on NAS-bench-201, and demonstrate the power of ABS by combining with various NAS algorithms on multiple search spaces and datasets. All experiments prove that the proposed method is highly efficient, and can significantly improve existing NAS algorithms.

However, there are some problems not solved yet.
For example, how to discriminate average pooling and max pooling; how to process more non-parametric operators such as different activation functions~\cite{glorot2011deep,maas2013rectifier,ramachandran2017searching}).
In the future, we will spend more time on discriminating more non-parametric operators using the angle-based metric in NAS. Additionally, we plan to apply our proposed metric to some downstream tasks (e.g., detection, segmentation) in our future work.
\clearpage
%
%
\bibliographystyle{splncs04}
\bibliography{egbib}

\begin{thebibliography}{10}
\providecommand{\url}[1]{\texttt{#1}}
\providecommand{\urlprefix}{URL }
\providecommand{\doi}[1]{https://doi.org/#1}

\bibitem{adam2019understanding}
Adam, G., Lorraine, J.: Understanding neural architecture search techniques.
  arXiv preprint arXiv:1904.00438  (2019)

\bibitem{arora2018theoretical}
Arora, S., Li, Z., Lyu, K.: Theoretical analysis of auto rate-tuning by batch
  normalization. arXiv preprint arXiv:1812.03981  (2018)

\bibitem{bender2018understanding}
Bender, G., Kindermans, P.J., Zoph, B., Vasudevan, V., Le, Q.: Understanding
  and simplifying one-shot architecture search. In: International Conference on
  Machine Learning. pp. 549--558 (2018)

\bibitem{cai2019once}
Cai, H., Gan, C., Han, S.: Once for all: Train one network and specialize it
  for efficient deployment. arXiv preprint arXiv:1908.09791  (2019)

\bibitem{cai2018proxylessnas}
Cai, H., Zhu, L., Han, S.: Proxylessnas: Direct neural architecture search on
  target task and hardware. arXiv preprint arXiv:1812.00332  (2018)

\bibitem{carbonnelle2019layer}
Carbonnelle, S., De~Vleeschouwer, C.: Layer rotation: a surprisingly simple
  indicator of generalization in deep networks?  (2019)

\bibitem{chen2019progressive}
Chen, X., Xie, L., Wu, J., Tian, Q.: Progressive differentiable architecture
  search: Bridging the depth gap between search and evaluation. arXiv preprint
  arXiv:1904.12760  (2019)

\bibitem{chen2019detnas}
Chen, Y., Yang, T., Zhang, X., Meng, G., Xiao, X., Sun, J.: Detnas: Backbone
  search for object detection. In: Advances in Neural Information Processing
  Systems. pp. 6638--6648 (2019)

\bibitem{chu2019fairnas}
Chu, X., Zhang, B., Xu, R., Li, J.: Fairnas: Rethinking evaluation fairness of
  weight sharing neural architecture search. arXiv preprint arXiv:1907.01845
  (2019)

\bibitem{dong2019one}
Dong, X., Yang, Y.: One-shot neural architecture search via self-evaluated
  template network. In: Proceedings of the IEEE International Conference on
  Computer Vision (ICCV). pp. 3681--3690 (2019)

\bibitem{dong2019searching}
Dong, X., Yang, Y.: Searching for a robust neural architecture in four gpu
  hours. In: Proceedings of the IEEE Conference on Computer Vision and Pattern
  Recognition. pp. 1761--1770 (2019)

\bibitem{dong2020nasbench201}
Dong, X., Yang, Y.: Nas-bench-201: Extending the scope of reproducible neural
  architecture search. In: International Conference on Learning Representations
  (ICLR) (2020), \url{https://openreview.net/forum?id=HJxyZkBKDr}

\bibitem{glorot2010understanding}
{Glorot}, X., {Bengio}, Y.: Understanding the difficulty of training deep
  feedforward neural networks. In: Proceedings of the Thirteenth International
  Conference on Artificial Intelligence and Statistics. pp. 249--256 (2010)

\bibitem{glorot2011deep}
Glorot, X., Bordes, A., Bengio, Y.: Deep sparse rectifier neural networks. In:
  Proceedings of the fourteenth international conference on artificial
  intelligence and statistics. pp. 315--323 (2011)

\bibitem{guo2019single}
Guo, Z., Zhang, X., Mu, H., Heng, W., Liu, Z., Wei, Y., Sun, J.: Single path
  one-shot neural architecture search with uniform sampling. arXiv preprint
  arXiv:1904.00420  (2019)

\bibitem{he2015delving}
{He}, K., {Zhang}, X., {Ren}, S., {Sun}, J.: Delving deep into rectifiers:
  Surpassing human-level performance on imagenet classification. In: 2015 IEEE
  International Conference on Computer Vision (ICCV). pp. 1026--1034 (2015)

\bibitem{ioffe2015batch}
Ioffe, S., Szegedy, C.: Batch normalization: Accelerating deep network training
  by reducing internal covariate shift. arXiv preprint arXiv:1502.03167  (2015)

\bibitem{kendall1938new}
Kendall, M.G.: A new measure of rank correlation. Biometrika  \textbf{30}(1/2),
   81--93 (1938)

\bibitem{krizhevsky2017imagenet}
{Krizhevsky}, A., {Sutskever}, I., {Hinton}, G.E.: Imagenet classification with
  deep convolutional neural networks. Communications of The ACM
  \textbf{60}(6),  84--90 (2017)

\bibitem{li2019improving}
Li, X., Lin, C., Li, C., Sun, M., Wu, W., Yan, J., Ouyang, W.: Improving
  one-shot nas by suppressing the posterior fading. arXiv preprint
  arXiv:1910.02543  (2019)

\bibitem{li2019exponential}
Li, Z., Arora, S.: An exponential learning rate schedule for deep learning.
  arXiv preprint arXiv:1910.07454  (2019)

\bibitem{liu2019auto}
Liu, C., Chen, L.C., Schroff, F., Adam, H., Hua, W., Yuille, A.L., Fei-Fei, L.:
  Auto-deeplab: Hierarchical neural architecture search for semantic image
  segmentation. In: Proceedings of the IEEE Conference on Computer Vision and
  Pattern Recognition. pp. 82--92 (2019)

\bibitem{liu2018progressive}
Liu, C., Zoph, B., Neumann, M., Shlens, J., Hua, W., Li, L.J., Fei-Fei, L.,
  Yuille, A., Huang, J., Murphy, K.: Progressive neural architecture search.
  In: Proceedings of the European Conference on Computer Vision (ECCV). pp.
  19--34 (2018)

\bibitem{liu2018darts}
Liu, H., Simonyan, K., Yang, Y.: Darts: Differentiable architecture search.
  arXiv preprint arXiv:1806.09055  (2018)

\bibitem{maas2013rectifier}
Maas, A.L., Hannun, A.Y., Ng, A.Y.: Rectifier nonlinearities improve neural
  network acoustic models. In: Proc. icml. vol.~30, p.~3 (2013)

\bibitem{nayman2019xnas}
Nayman, N., Noy, A., Ridnik, T., Friedman, I., Jin, R., Zelnik, L.: Xnas:
  Neural architecture search with expert advice. In: Advances in Neural
  Information Processing Systems. pp. 1975--1985 (2019)

\bibitem{noy2019asap}
Noy, A., Nayman, N., Ridnik, T., Zamir, N., Doveh, S., Friedman, I., Giryes,
  R., Zelnik-Manor, L.: Asap: Architecture search, anneal and prune. arXiv
  preprint arXiv:1904.04123  (2019)

\bibitem{perez2018efficient}
P{\'e}rez-R{\'u}a, J.M., Baccouche, M., Pateux, S.: Efficient progressive
  neural architecture search. arXiv preprint arXiv:1808.00391  (2018)

\bibitem{pham2018efficient}
Pham, H., Guan, M.Y., Zoph, B., Le, Q.V., Dean, J.: Efficient neural
  architecture search via parameter sharing. arXiv preprint arXiv:1802.03268
  (2018)

\bibitem{ramachandran2017searching}
Ramachandran, P., Zoph, B., Le, Q.V.: Searching for activation functions. arXiv
  preprint arXiv:1710.05941  (2017)

\bibitem{wan2020spherical}
{Wan}, R., {Zhu}, Z., {Zhang}, X., {Sun}, J.: Spherical motion dynamics of deep
  neural networks with batch normalization and weight decay. arXiv preprint
  arXiv:2006.08419  (2020)

\bibitem{wang2019scalable}
Wang, L., Xie, L., Zhang, T., Guo, J., Tian, Q.: Scalable nas with factorizable
  architectural parameters. arXiv preprint arXiv:1912.13256  (2019)

\bibitem{wu2019fbnet}
Wu, B., Dai, X., Zhang, P., Wang, Y., Sun, F., Wu, Y., Tian, Y., Vajda, P.,
  Jia, Y., Keutzer, K.: Fbnet: Hardware-aware efficient convnet design via
  differentiable neural architecture search. In: Proceedings of the IEEE
  Conference on Computer Vision and Pattern Recognition. pp. 10734--10742
  (2019)

\bibitem{xu2019auto}
Xu, H., Yao, L., Zhang, W., Liang, X., Li, Z.: Auto-fpn: Automatic network
  architecture adaptation for object detection beyond classification. In:
  Proceedings of the IEEE International Conference on Computer Vision. pp.
  6649--6658 (2019)

\bibitem{zhang2020deeper}
Zhang, Y., Lin, Z., Jiang, J., Zhang, Q., Wang, Y., Xue, H., Zhang, C., Yang,
  Y.: Deeper insights into weight sharing in neural architecture search. arXiv
  preprint arXiv:2001.01431  (2020)

\bibitem{zoph2018learning}
Zoph, B., Vasudevan, V., Shlens, J., Le, Q.V.: Learning transferable
  architectures for scalable image recognition. In: Proceedings of the IEEE
  conference on computer vision and pattern recognition. pp. 8697--8710 (2018)

\end{thebibliography}

\newpage
\appendix
\setcounter{secnumdepth}{0}
\setcounter{figure}{0}
\setcounter{table}{0}
\section{Appendix}

\begin{figure}[h]
\centering
 \includegraphics[width=1.0\linewidth]{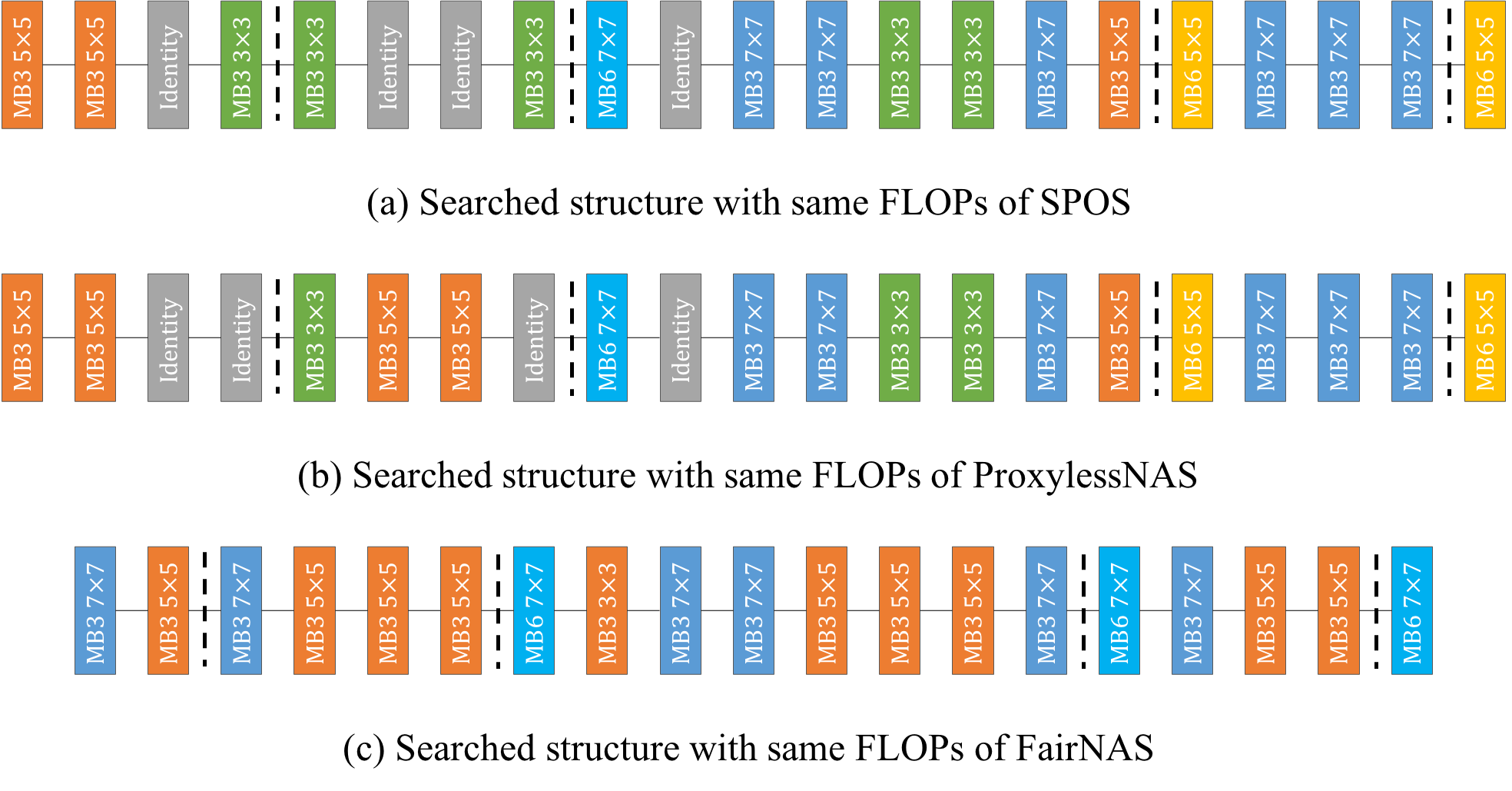}
\centering
\caption{Structures of searched architectures under FLOPs constraints over the MobileNet-like shrunk search space, see Table 4 for details
}
\label{fig:structures_proxylessNAS_search_space}
\end{figure}

\begin{table}[h]
\caption{The original search space S$1$ and its shrunk search spaces of various size from NAS-Bench-201, see Section 3.1 for details}
\label{table:search space benchmark}
\begin{center}
\begin{tabular}{cc}
\hline\
Search Space & Candidate Operators\\
\hline
    S1 & $none$, $skip$ $connect$, $conv$ $1\times1$, $conv$ $3\times3$, $average$ $pooling$ $3\times3$\\
    S2 & $skip$ $connect$, $conv$ $1\times1$, $conv$ $3\times3$, $average$ $pooling$ $3\times3$\\
    S3 & $none$, $conv$ $1\times1$, $conv$ $3\times3$, $average$ $pooling$ $3\times3$\\
    
    S4 & $none$, $skip$ $connect$, $conv$ $1\times1$, $average$ $pooling$ $3\times3$\\
    S5 & $none$, $skip$ $connect$, $conv$ $1\times1$, $conv$ $3\times3$\\
    S6 & $conv$ $1\times1$, $conv$ $3\times3$, $average$ $pooling$ $3\times3$\\
    S7 & $none$, $skip$ $connect$, $average$ $pooling$ $3\times3$\\
    S8 & $conv$ $1\times1$, $conv$ $3\times3$\\
\hline
\end{tabular}
\end{center}
\end{table}

\begin{figure}[t]
\centering
\subfigure[Normal cell searched by PDARTS on CIFAR-10]{
        \includegraphics[width=\linewidth]{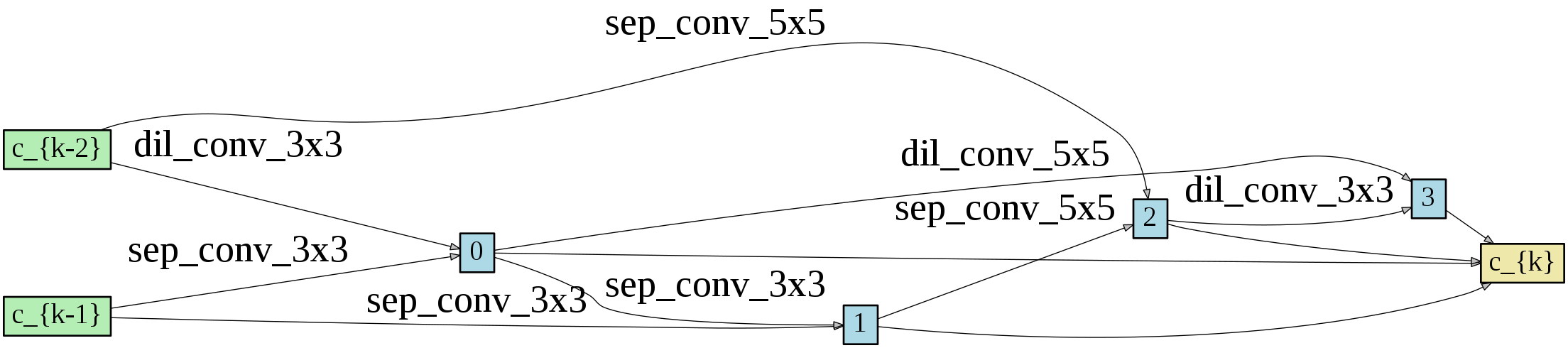}
}
\subfigure[Reduction cell searched by PDARTS on CIFAR-10]{
    \includegraphics[width=\linewidth]{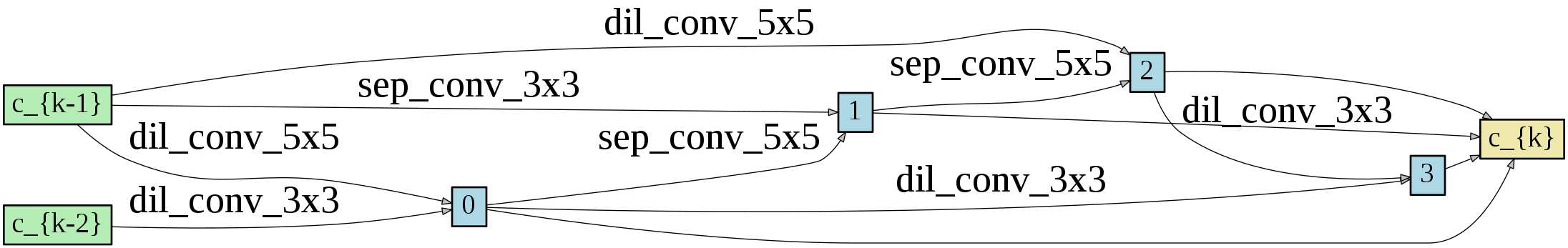}
}
    \subfigure[Normal cell searched by DARTS on CIFAR-10]{
        \includegraphics[width=\linewidth]{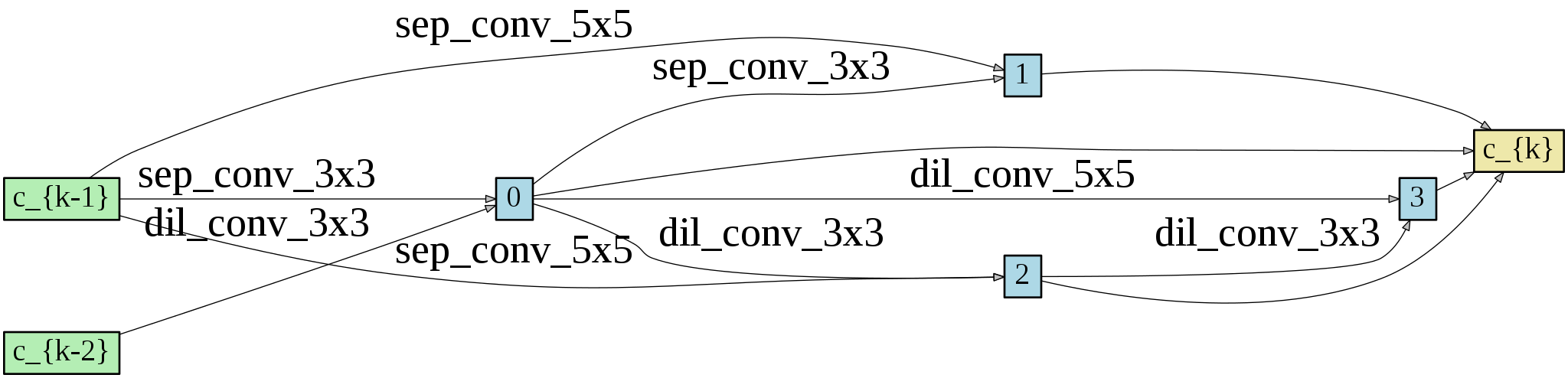}
    }
    \subfigure[Reduction cell searched by DARTS on CIFAR-10]{
        \includegraphics[width=0.85\linewidth]{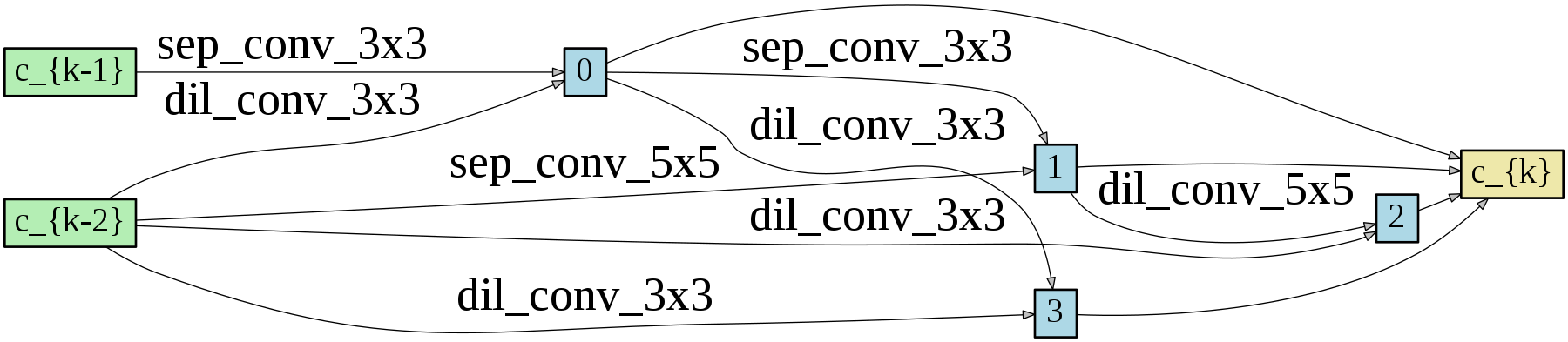}
    }
\centering
\caption{Structures of searched architectures over the shrunk search space of DARTS, see Table 5 and 6 for details}
\label{fig:structure_pdarts}
\end{figure}

\begin{figure}[h]
\centering
 \includegraphics[width=1.0\linewidth]{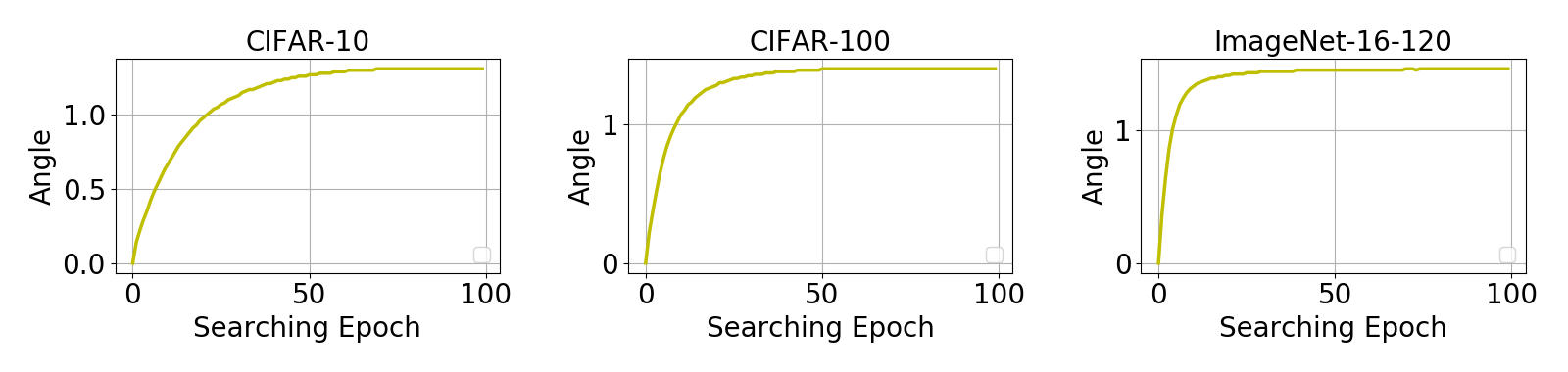}
\centering
\caption{The angle evolution of a standalone model on different datasets. The model is chosen from NAS-Benchmark-201 search space. The angle values adopt radian measure.}
\label{fig:angle_evolution_between_epochs}
\end{figure}

\begin{figure}[h]
\centering
 \includegraphics[width=1.0\linewidth]{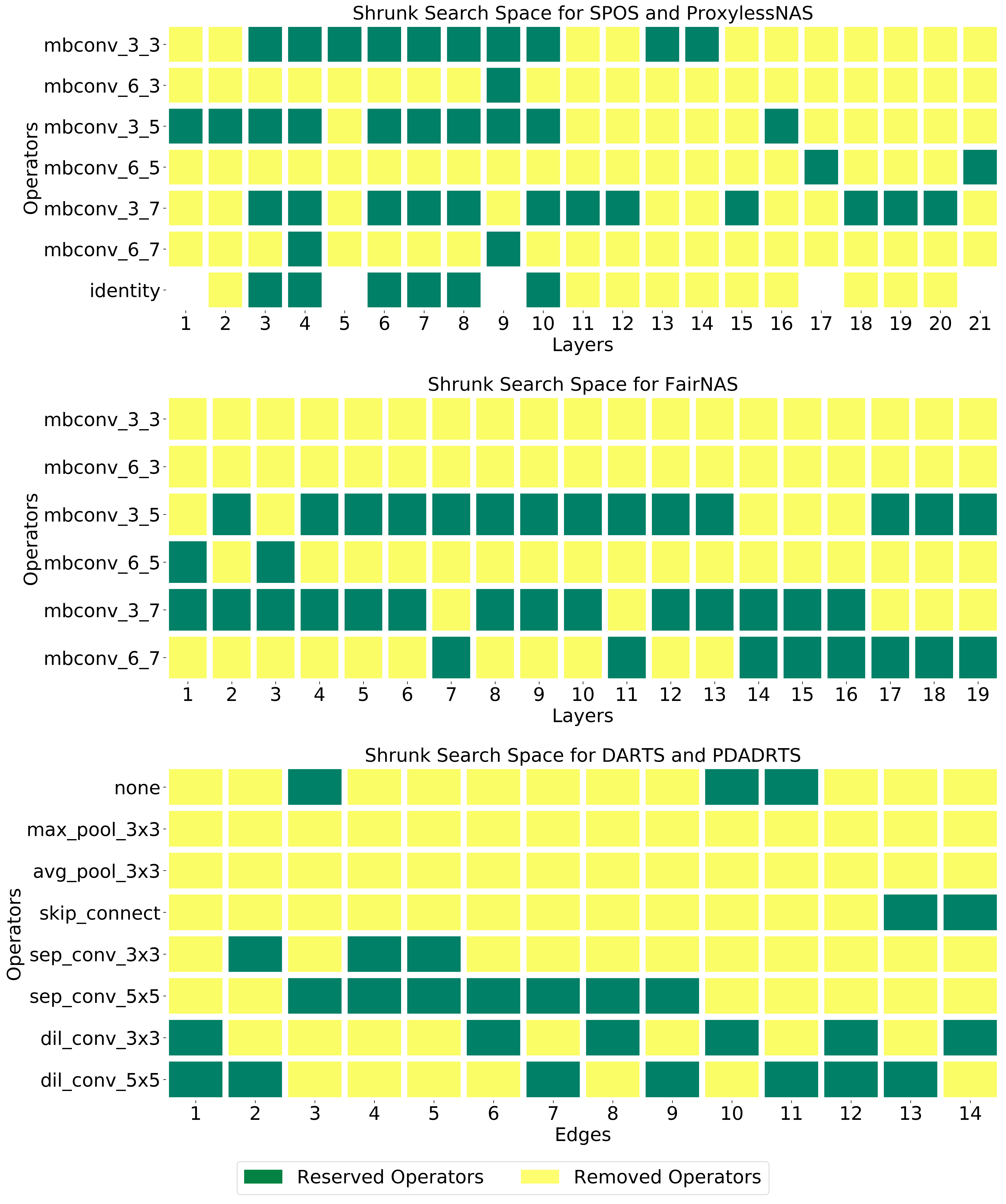}
\centering
\caption{Shrunk search spaces found by ABS. MobileNet-like search space contains a set of $mbconv$ operators (mobile inverted bottleneck convolution), where $mbconv\_X\_Y$ represents the specific operator with expand ratio $X$ and kernel size $Y$. See Tablets 4, 5, 6 for details}
\label{fig:shrunk search spaces}
\end{figure}

\begin{table}[h]
\caption{Comparison between the best search space `S8' designed by human and the one found by ABS on NASBench-201. For the form x(y), x means represents the searched models on `S8', y represents the searched models on the shrunk search space ABS finds}
\label{table:comparison_ABS_human}
\begin{center}
\begin{tabular}{cccccc}
\hline
Method & CIFAR-10 & CIFAR-100 & ImageNet-16-120\\
\hline
DARTS \cite{liu2018darts} & 89.37\% (88.90\%) & 65.70\% (67.57\%) & 35.47\% (33.19\%) \\
SETN \cite{dong2019one} & 93.20\% (93.07\%) & 70.12\% (69.12\%) & 44.23\% (46.33\%) \\ 
ENAS \cite{pham2018efficient} & 93.76\% (93.76\%) & 71.10\% (71.10\%) & 35.47\% (41.44\%) \\
GDAS \cite{dong2019searching} & 93.47\% (93.60\%) & 71.01\% (70.35\%) & 44.89\% (41.02\%) \\
SPOS \cite{guo2019single} & 93.79\% (93.76\%) & 70.49\% (71.95\%) & 44.68\% (44.24\%) \\
\hline
\end{tabular}
\end{center}
\end{table}

\begin{figure}[h]
\centering
 \includegraphics[width=1.0\linewidth]{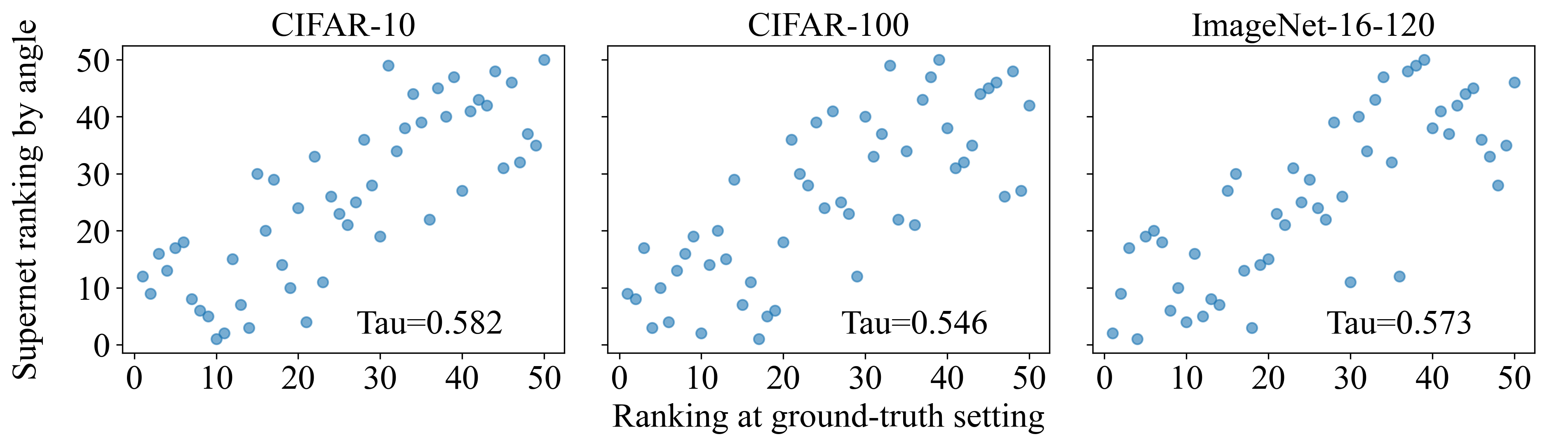}
\centering
\caption{The distribution of the supernet ranking by the angle-based metric. We uniformly choose 50 models from NAS-Bench-201, and rank them based on supernet. X-axis represents the ground-truth ranking based on the ground-truth accuracy}
\label{fig:shrunk search spaces}
\end{figure}

\begin{table}[h]
\caption{The removed operators for each shrinking iteration on MobileNet-like search space. For the form (x,y), x denotes layer index, y represents operator type. Identity, MB3\_3$\times$3, MB6\_3$\times$3, MB3\_5$\times$5, MB6\_5$\times$5, MB3\_7$\times$7 and MB6\_7$\times$7 are encoded from -1 to 5 respectively }
\label{table:shrinking_procedure}
\begin{center}
\begin{tabular}{cc}
\hline
Iteration & Removed Operators\\
\hline
1  & (19, 1), (17, 1), (17, 3), (18, 1), (18, 3), (19, 5), (19, 3)\\
2 & (18, 5), (17, 5), (13, 3), (14, 1), (15, 3), (14, 3), (14, 5)\\ 
3 & (15, 1), (13, 1), (15, 5), (9, 1), (10, 1), (10, 3), (11, 3)\\ 
4 & (11, 5), (9, 3), (9, 5), (5, 3), (20, 5), (13, 5), (7, 1)\\
5 & (7, 5), (12, 1), (6, 3), (3, 1), (10, 5), (6, 1), (6, 5)\\
6 & (2, 1), (5, 5), (16, 5), (16, 1), (2, 3), (3, 3), (11, 1)\\
7 & (1, 3), (1, 1), (7, 3), (1, 5), (8, 3), (2, 5), (1, 0)\\
8 & (4, 1), (12, 3), (0, 0), (5, 1), (17, 2), (4, 4), (0, 1)\\
9 & (17, -1), (18, -1), (19, -1), (20, 0), (1, 4), (1, -1), (15, -1)\\
10 & (17, 0), (20, 2), (20, 4), (19, 0), (18, 0), (19, 2), (15, 4)\\
11 & (20, 1), (18, 2), (15, 0), (13, -1), (12, 4), (10, -1), (11, -1)\\
12 & (12, 2), (0, 4), (16, 4), (12, 5), (14, -1), (10, 0), (4, 2)\\
13 & (4, 5), (0, 5), (11, 0), (14, 2), (16, 2), (13, 4), (13, 2)\\
14 & (4, 3), (11, 2), (0, 3), (10, 2), (14, 0), (16, 0), (8, 4)\\
\hline
\end{tabular}
\end{center}
\end{table}

\end{document}